# Validating Emergency Department Admission Predictions Based on Local Data Through MIMIC-IV

Francesca Meimeti [1], Loukas Triantafyllopoulos [1*], Aikaterini Sakagianni [2], Vasileios Kaldis [3], Lazaros Tzelves [4], Nikolaos Theodorakis [5,6], Evgenia Paxinou[1], Georgios Feretzakis[1], Dimitris Kalles[1] and Vassilios S. Verykios [1*]

[1] School of Science and Technology, Hellenic Open University, Patras, Greece; francesca.meimeti@gmail.com (F.M.); triantafillopoulos.loukas@ac.eap.gr (L.Tr.); paxinou.evgenia@ac.eap.gr (E.P.); georgios.feretzakis@ac.eap.gr (G.F.); kalles@eap.gr (D.K.); verykios@eap.gr (V.V.)

[2] Intensive Care Unit, Sismanogleio General Hospital, 15126 Marousi, Greece; sakagianni@sismanoglio.gr (A.S.)

[3] Emergency Department, Sismanogleio General Hospital, 15126 Marousi Greece; dnths-tep@sismanoglio.gr, vkaldis@yahoo.gr (V.K.)

[4] 2nd Department of Urology, Sismanoglio General Hospital, Sismanogliou 37, National and Kapodistrian University of Athens, 15126 Athens, Greece; lazarostzelves@gmail.com (L.Tz.)

[5] Department of Cardiology & 65+ Clinic, Amalia Fleming General Hospital, 14, 25th Martiou Str., School of Medicine, National and Kapodistrian University of Athens, 75 Mikras Asias Str., 11527 Athens, Greece; nikolaostheodorakis1997@yahoo.com (N.T.)

*Correspondence: triantafillopoulos.loukas@ac.eap.gr; verykios@eap.gr

**Abstract**

The effective management of Emergency Department (ED) overcrowding is essential for improving patient outcomes and optimizing healthcare resource allocation. This study validates hospital admission prediction models initially developed using a small local dataset from a Greek hospital by leveraging the comprehensive MIMIC-IV dataset. After preprocessing the MIMIC-IV data, five algorithms—Linear Discriminant Analysis (LDA), K-Nearest Neighbors (KNN), Random Forest (RF), Recursive Partitioning and Regression Trees (RPART), and Support Vector Machines (svmRadial)—were evaluated. Among these, RF demonstrated superior performance, achieving an Area Under the Receiver Operating Characteristic Curve (AUC-ROC) of 0.9999, sensitivity of 0.9997, and specificity of 0.9999 when applied to the MIMIC-IV data. These findings underscore the robustness of RF in handling complex datasets for admission prediction, establishing MIMIC-IV as a valuable benchmark for validating models based on smaller local datasets, and providing actionable insights for steering ED management strategies in the right direction.

# 1. Introduction

A key recent issue in healthcare is the overcrowding of emergency departments (EDs) [1-2]. In developed countries with rapidly ageing populations, the role of EDs is particularly important [3]. At the same time, this issue often requires an immediate and coordinated response at a global level, especially in situations where the extent of the dire consequences of a health crisis may be difficult to predict [4].

Given that every ED has a limited number of human and non-human resources, excessive patient admissions have been shown to lead to a variety of negative outcomes [5]. This occurs because, when EDs are overcrowded, the quality of care provided declines, patients may experience significant delays, and some may even leave the hospital without being examined [6]. As a result, the same patients are likely to return to EDs later with more severe illnesses, leading to prolonged hospital stays, and higher mortality rates in these units [7]. The negative implications, in fact, may even extend to the emotional part, both for patients and staff, with increased patient dissatisfaction and decreased staff morale [8].

As a culmination of efforts to address this problematic situation, predictive models are being promoted in EDs [9]. These models aim to specifically predict the volume of patients that an ED can serve, identify patients considered to be at high risk, and prepare ED resources effectively for situations of intense demand and overcrowding [10]. They are designed to mitigate the risk of human error during the decision-making process, which is particularly important in emergency care, where time is limited, and mistakes can cost lives [11].

A prerequisite for a positive outcome is, of course, the accuracy of these models [12]. Their reliability largely depends on the quality and sensitivity of the input data, which can vary significantly across different healthcare settings [13]. ED prediction models utilize a variety of data, including triage information, vital signs, demographic details, medications, and chief complaints [14]. However, the availability of these data varies across national datasets [15]. Some datasets may lack granularity or detail on clinical variables, which could be critical for accurately understanding patient admissions in the ED. This raises questions about the validity and accuracy of the predictive models they support [14, 15].

An effective alternative for addressing data limitations in ED predictive models is the use of the Medical Information Mart for Intensive Care (MIMIC) database [16], a comprehensive and publicly accessible resource developed through a collaboration between Beth Israel Deaconess Medical Center (BIDMC) and the Massachusetts Institute of Technology (MIT) [17]. This free database, currently in its fourth version, contains anonymized data from BIDMC's intensive care unit, encompassing extensive and detailed information gathered during routine clinical practice [18]. Unlike typical hospital archival systems, which are optimized for storage and often restrict researcher access, MIMIC is structured specifically to support research. It includes key patient demographics, admission diagnoses, therapeutic profiles, orders, procedures, treatments, and de-identified clinical notes, offering valuable insights into patient care and treatment responses while enabling studies aimed at improving patient outcomes [19].

This database is considered a promising tool for enhancing ED admission prediction accuracy, with significant implications for healthcare policy and resource allocation [13, 20]. Its use could help optimize hospital resources, alleviate ED crowding, and improve patient outcomes by enabling early notification of administrators and inpatient teams for better planning and response [14].

In this context, the present study focuses on comparing and validating findings related to the prediction of patient admissions to the ED of a Greek public hospital with those obtained using the MIMIC-IV database. The structure of this paper is organized as follows: Section 2 reviews the relevant literature and outlines the specific research questions addressed by this study. Section 3 describes our methodology, providing an overview of the datasets, the preprocessing steps applied, the algorithms implemented, and the evaluation metrics used to compare the Greek local dataset with the MIMIC-IV dataset. Section 4 presents the empirical findings of the comparative analysis. Sections 5 and 6 discuss the results, implications, and limitations of the study, concluding with key insights and recommendations for future research, respectively.

## 2. Background and literature

Within the context of the relevant literature, this section focuses on two important areas. First, it reviews studies that utilize national or local databases to predict ED admissions, exploring the advantages and limitations of these large-scale, population-based datasets. Secondly, it examines research employing the MIMIC database for predictive modelling in critical care and emergency settings, highlighting the value of MIMIC's detailed clinical data and its applicability to high-risk patient scenarios.

*2.1. Prediction Models for ED Outcomes Using National or Local Health Data*

Considering recent literature, several studies have used remarkable local hospital data from around the world to predict patient admissions to EDs. Notably, intense re-search activity is observed in the U.S.A., where most studies on this scientific issue are concentrated.

Specifically, in Massachusetts, Subudhi et al. [21] applied Random Forest models to predict Intensive Care Unit (ICU) admissions for 3,597 COVID-19 patients, achieving an F1 score of 0.810. Similarly, Douda et al. [22] in Michigan focused on ICU admissions for 1,094 COVID-19 patients, using logistic regression to achieve a c-statistic of 0.798 in derivation and 0.764 in validation, emphasizing the potential of predictive models for crisis resource management. Around the same time as Subudhi's work, Fenn et al. [23] in North Carolina used data from 468,167 ED encounters to develop models with Area Under the Receiver Operating Characteristic Curve (AUC-ROC) of 0.951 for ICU ad-missions. While Nguyen et al. [24], at Stanford University, developed a gradient-boosted tree model for ICU predictions, achieving an AUC-ROC of 0.880, which Ip et al. [25] later expanded with a multimodal Artificial Intelligence (AI) model combining video and triage data, achieving an AUC-ROC of 0.714. Furthermore, Ahmed et al. [26] introduced in the Midwest the T-ADAB model using 453,664 ED records, optimized with Tabu Search to achieve an AUC-ROC of 95.4% with 99.3% sensitivity and 97.2% accuracy. Also, Monahan et al. in Alabama used logistic regression on triage data from 93,847 ED patients, achieving an AUC-ROC of 0.841 [27]. Finally, in Pennsylvania, Pai et al. [28] applied neural networks to predict admissions for fall-related fractures, achieving an AUC-ROC of 0.938 with triage data alone and 0.983 when incorporating post-diagnosis information. And, last year Glicksberg et al. [29] in New York City applied GPT-4 with retrieval-augmented generation (RAG) to data from seven hospitals, achieving an AUC-ROC of 0.870 and an accuracy of 83.1%.

Across Europe, machine learning (ML) models have advanced emergency admission predictions using diverse datasets and methods. In Scotland, Liley et al. [30] developed SPARRAv4, achieving an AUC-ROC of 0.799 with national health records from 4.8 million residents, demonstrating improved calibration and stability over three years. In Greece, Feretzakis et al. showcased the utility of random forest models for predicting ad-missions from 13,991 visits [31]. Furthermore, in the UK, Stylianou et al. [32] achieved an AUC-ROC of 0.9384 with logistic regression on data from 190,466 individuals, while King et al. [33] employed XGBoost on 109,465 EHR records, achieving AUC-ROCs of 0.82–0.90 depending on visit duration. Similarly, in Spain, Cusidó et al. [34] developed a gradient boosting machine model with an AUC-ROC of 0.8938 using data from 3.1 million ED visits, and Álvarez-Chaves et al. [35] enhanced forecasting accuracy with GAN-augmented models, reducing SMAPE to 6.99 [35]. Additionally, in the Nether-lands, De Hond et al. [36] achieved AUC-ROCs of up to 0.860 using triage data from 172,104 visits. Meanwhile, in France, Brossard et al. [37] optimized XGBoost models, achieving low mean absolute errors across 87,600 time slots. Finally, Leonard et al. [38] in Ireland developed a gradient boosting model with an AUC-ROC of 0.835 using pediatric ED data, offering efficient predictions in resource-limited settings.

Noteworthy research activity in other countries includes several advances in predictive modeling in Australia. Pandey et al. [39] and Kishore et al. [40] at Austin Hospital developed ML models using ED data to predict ICU and hospital admissions, achieving AUC-ROCs of 0.920 and ≥0.930, respectively, with notable improvements in accuracy and efficiency. Building on the use of ED data, Higgins et al. (2024) in New South Wales applied explainable AI to triage data, successfully predicting acute mental health ward ad-missions [41]. Expanding these efforts to other regions, Lee et al. (2021) in Taiwan utilized a neural network to predict urgent ED admissions, achieving an AUC-ROC of 0.800 [42]. Similarly, Zahid et al. (2023) in Qatar introduced the MAPS tool, which leveraged triage data from 320,299 visits to achieve an AUC-ROC of 0.831 and 83.3% accuracy [43]. Lastly, adding to this global body of work, Ratnovsky et al. [44] in Israel applied an Artificial Neural Network (ANN) model to 124,915 ED visits, achieving an AUC-ROC of 0.790. A summary of the key findings from a number of comparable studies is provided in Table 1, highlighting the models, sample sizes, and performance metrics used across different regions.

Despite extensive research on ED admission prediction worldwide, significant variations exist in the sizes of training datasets, methodologies, and predictors employed across studies. Relying on local data —whether from a single hospital or a specific national healthcare system—may limit generalizability to other healthcare settings. This highlights the need for a robust framework to test and validate the predictive performance of local data using an internationally recognized dataset. The MIMIC database, with its comprehensive clinical records and standardized structure, offers an ideal platform for applying ML techniques to enhance the predictive accuracy of ED outcomes across diverse patient populations.

**Table 1.** Summary of Predictive Modeling Studies on ED Admissions Worldwide

| Study (Year) | Location | Sample Size | Model Used | Performance Metric | Score |
|---|---|---|---|---|---|
| Subudhi et al. (2021) | Massachusetts, USA | 3,597 | AdaBoostClassifier | F1 score | 0.75–0.85 |
| | | | BaggingClassifier | | 0.77–0.85 |
| | | | GradientBoostingClassifier | | 0.77–0.85 |
| | | | RandomForestClassifier | | 0.78–0.84 |
| | | | XGBClassifier | | 0.76–0.84 |
| | | | ExtraTreesClassifier | | 0.76–0.84 |
| | | | LogisticRegression | | 0.73–0.81 |
| | | | DecisionTreeClassifier | | 0.76–0.80 |
| | | | LinearDiscriminantAnalysis | | 0.72–0.82 |
| | | | QuadraticDiscriminantAnalysis | | 0.78–0.80 |
| Douda et al. (2024) | Michigan, USA | 1,094 | Logistic Regression | AU-ROC | 0.764 |
| | | | | Sensitivity | 0.721 |
| | | | | Specificity | 0.763 |
| Fenn et al. (2021) | Durham, USA | 468,167 | Gradient-Boosted Trees (LightGBM) | AU-ROC | 0.873 |
| Nguyen et al. (2021) | Stanford, USA | 41,654 | Gradient-Boosted Trees | AU-ROC | 0.88 |
| | | | Logistic Regression (ESI-based) | | 0.67 |
| | | | Random Forest | | 0.86 |
| Ip et al. (2024) | Palo Alto, USA | 723 | Baseline ESI Model | AU-ROC | 0.575 |
| | | | Triage Data Only Model | | 0.678 |
| | | | Video Data Only Model | | 0.693 |
| | | | Late Fusion Model (Video+Triage) | | 0.714 |
| Ahmed et al. (2022) | Midwest, USA | 5,000 (out of 400k) | Optimized AdaBoost | AU-ROC | 0.954 |
| | | | Optimized XGBoost | | 0.948 |
| | | | Optimized MLP | | 0.879 |
| | | | Traditional XGBoost | | 0.9040 - 0.9410 |
| | | | Traditional AdaBoost | | 0.8820 - 0.9320 |
| | | | Traditional Multilayer Perceptron | | 0.8080 - 0.8740 |
| Glicksberg et al. (2024) | New York City, USA | 864,089 | Ensemble (Bio-Clinical-BERT + XGBoost) | AU-ROC | 0.878 |
| | | | | Accuracy | 0.829 |
| Monahan et al. (2023) | Alabama, United States | 113,739 | Multivariable Fractional Polynomial Logistic Regression | AU-ROC | 0.841 |
| | | | 10-Fold Cross-Validated MFP Logistic Regression | | 0.839-0.842 |

**Table 1.** Summary of Predictive Modeling Studies on ED Admissions Worldwide (continue)

| Study (Year) | Location | Sample Size | Model Used | Performance Metric | Score |
|---|---|---|---|---|---|
| Pai et al. (2023) | Pennsylvania, USA | 6,335 | Neural Network | AU-ROC | 0.983 |
| | | | Logistic Regression | | 0.936 |
| | | | k-Nearest Neighbors | | 0.936 |
| Liley et al. (2024) | Scotland | 4,800,000 | SPARRAv4 | AU-ROC | 0.799 |
| | | | XGBoost | | 0.798 |
| | | | Random Forest | | 0.792 |
| | | | Logistic Regression | | 0.788 |
| | | | Naive Bayes | | 0.747 |
| Feretzakis et al. (2022) | Athens, Greece | 13,991 | Random Forest | AUC – ROC \| Sensitivity \| Specificity | 0.805 \| 0.697 \| 0.776 |
| | | | Support Vector Machines | AUC – ROC \| Sensitivity \| Specificity | 0.796 \| 0.701 \| 0.769 |
| | | | k-Nearest Neighbors | AUC – ROC \| Sensitivity \| Specificity | 0.731 \| 0.678 \| 0.680 |
| | | | Linear Discriminant Analysis | AUC – ROC \| Sensitivity \| Specificity | 0.783 \| 0.717 \| 0.718 |
| | | | Recursive Partitioning and Regression Trees | AUC – ROC \| Sensitivity \| Specificity | 0.699 \| 0.583 \| 0.762 |
| Stylianou et al. (2022) | Bath, UK | 190,466 | Logistic Regression | AU-ROC | 0.9384 |
| King et al. (2022) | London, UK | 109,465 | XGBoost Classifier | AU-ROC | 0.82 - 0.90 |
| Alvarez-Chaves et al. (2024) | Madrid, Spain | 361,698 | Generative AI (DoppelGANger) with Prophet Model | SMAPE | 6.79 - 8.56 |
| De Hond et al. (2021) | Netherlands | 172,104 | Logistic Regression | AU-ROC | 0.83 |
| | | | Random Forest | | 0.86 |
| | | | Gradient Boosted Decision Trees (XGBoost) | | 0.86 |
| | | | Deep Neural Network | | 0.86 |
| Leonard et al. (2022) | Dublin, Ireland | 76,000 | Logistic Regression, Naïve Bayes, Gradient Boosting Machine | AU-ROC | 0.789-0.913 |
| Pandey et al. (2024) | Melbourne, Australia | 484,094 | Natural Language Processing | AU-ROC | 0.921 |
| Kishore et al. (2023) | Melbourne, Australia | 599,015 | Machine Learning | AU-ROC | ≥0.93 |
| | | | | Sensitivity | 0.83 |
| | | | | Specificity | 0.90 |
| | | | | F1-score | 0.84 |
| Lee et al. (2021) | Tainan, Taiwan | 282,971 | Neural network | AU-ROC | 0.8004 |
| Zahid et al. (2023) | Doha, Qatar | 320,299 | Logistic Regression (MAPS) | AU-ROC | 0.831 |
| | | | | Predictive Accuracy | 0.833 |
| | | | | Sensitivity | 0.691 |
| | | | | Specificity | 0.839 |
| Ratnovsky et al. (2021) | Tel Aviv, Israel | 124,915 | Artificial Neural Network (ANN) | AU-ROC | 0.79 |

*2.2. ED Admission Prediction Using the MIMIC Database*

From the literature review, it is evident that the MIMIC database could be an important tool for validating prediction results in EDs [45]. The breadth of available data makes this database an increasingly popular choice, especially with the highly upgraded latest version, MIMIC-IV.

Tsoni et al. [20] employed 280,000 records from the MIMIC-IV-ED database to predict hospital admissions upon a patient's arrival at the ED, emphasizing early-stage predictions based on basic vital signs. They tested various ML algorithms, with Gradient Boosted Trees delivering the best performance. Implemented through an open-source ML pipeline, their approach achieved 80% accuracy using only the initial triage data [20].

Similarly in Greece, Feretzakis et al. [13] leveraged AutoML with the MIMIC-IV-ED database to enhance ED triage by predicting hospital admissions. Using H2O.ai's AutoML, the Gradient Boosting Machine (GBM) model was developed using the same volume of records from BIDMC (2011–2019), yielding an AUC-ROC of 0.826. Key predictors included acuity and waiting time, showcasing the potential of automated approaches to refine triage processes [13].

Expanding on predictive modeling, Xie et al. [45] benchmarked ML models for predicting ED outcomes, including hospital admissions, using the MIMIC-IV-ED database. Their Gradient Boosting model was fine-tuned on data from 448,972 ED visits reaching an AUROC of 0.819, underlining the efficacy of ML in streamlining triage decisions and optimizing resource allocation in emergency settings [45].

In a novel approach, Kouhounestani et al. [46] developed TE-PrepNet, a preprocessing framework designed to improve hospital admission predictions at ED triage. Their Random Forest model, trained on the MIMIC-IV-ED database (440,285 instances), attained an AUC-ROC of 0.846, significantly outperforming the baseline model (AUC-ROC 0.752). Incorporating nominal features such as chief complaints and transport modes, TE-PrepNet demonstrated enhanced predictive accuracy and potential for optimizing triage decisions [46].

Shu et al. [47] focused on ICU admissions for patients with ischemic heart disease following ED visits. Using the MIMIC-IV database, their scoring model incorporated patient-level clinical parameters such as vital signs, lab results, and comorbidities. This model achieved high predictive accuracy, demonstrating its utility in identifying ED patients at risk of ICU transfer and optimizing resource allocation [47].

Further advancing ICU admission prediction, Choi et al. [48] at Korea University developed a Long Short-Term Memory Encoder-Decoder (LSTM-ED) model to predict ICU admissions among low-acuity triaged patients. Using 297,807 ED stays and 114 predictors from the MIMIC-IV database, their anomaly detection approach achieved a recall of 83%, precision of 82%, and an average lead time of 3.13 hours before ICU transfer. This model effectively identified 36% of all possible clinical deterioration events during a patient's ED stay, demonstrating its value in early detection of critical cases [48].

Lastly, El Ariss et al. [14] utilized records from the MIMIC-IV dataset, analyzing 391,472 patient visits. These records included vital signs, demographics, arrival mode, medication history, and chief complaints, which were used to train ML models for predicting resource needs. Their models achieved an average AUC-ROC of 0.820 and accuracy of 0.760, with chief complaints consistently ranking as a significant predictor, underscoring their importance in resource planning [14].

The aforementioned studies merely indicate that this particular database allows for the validation of various prognostic models for critical clinical outcomes in the ED, making it a useful tool. With the large volume of data and variables it contains, MIMIC-IV database could assist in evaluating the predictive results of small local studies.

*2.3. Research gap and questions*

After reviewing the recent literature and noting the abundance of studies that validate prediction models using a small number of local or national data, it seems particularly interesting to use the MIMIC database to confirm the results of these studies. This is something that has not been addressed in the literature so far, and constitutes a gap that the present study aims to fill. Specifically, it focuses on the following research questions:

1. Will we reach the same conclusions regarding the prediction of ED admissions based on a small set of local data when using the MIMIC database, employing the same ML algorithms and variables of interest?

2. Are there significant differences in forecasting accuracy between the study using a wealth of data from MIMIC and the study based on sparse local data, across the set of forecasting algorithms examined?

## 3. Methodology

To address our research questions, we used a study conducted in Greece by Feretzakis et al. [49] as a benchmark for comparing the MIMIC results in predicting ED admissions, given the established algorithms employed and the structured nature of the dataset. This study has one of the smallest local datasets compared to the other studies listed in Table 1. In this section, we present the datasets used, the preprocessing steps applied, the algorithms implemented, and the evaluation metrics employed in the two studies under comparison

### *3.1. Dataset Overview and Preparation for Comparative Analysis*

The local dataset used in the study by Feretzakis et al. [49] comprises 13,991 ED visits recorded at a Greek tertiary public hospital over a one-year period, from January to December 2020. It includes information on various biochemical markers such as Creatine Kinase (CPK), Creatinine (CREA), C-Reactive Protein (CRP), Lactate Dehydrogenase (LDH), serum Urea (UREA), activated partial thromboplastin time (aPTT), D-Dimer, International Normalized Ratio (INR), hemoglobin (HGB), lymphocyte count (LYM%), neutrophil count (NEUT%), platelets (PLT), and white blood cells (WBC). Additionally, the dataset contains patient demographics (age and gender), details about triage disposition in the ED (admission or discharge), arrival method (e.g., ambulance), and ED outcomes (admission, transfer, etc.). Additionally, patient data included age, gender, triage disposition in the ED (admission or discharge), arrival method (e.g., ambulance), and ED triage disposition (admission, transfer, etc.). From the mentioned information, the input variables were the medical markers (e.g., CPK, CREA, CRP, etc.) and demographic data (age, gender), while the output variable was the ED admission status or triage outcome. Thus, the prediction problem is framed as a binary classification task, where each ED visit is classified as either 'admitted' (Yes) or 'not admitted' (No).

The same data variables were extracted from the MIMIC-IV database (version 2.2) using SQLite software by an accredited member of this research group (the SQL code used is presented in Appendix Table A1). In total, we utilized a substantially larger sample size, consisting of 322,189 ICU-level hospitalization cases. The descriptive statistics of these raw data are presented in Table 2.

**Table 2.** Descriptive Statistics of the Initial Data Extracted from the MIMIC-IV

| Variable | Minimum Value | 1st Quartile | Median | Mean | 3rd Quartile | Maximum Value | Missing Values (NA's) |
|---|---|---|---|---|---|---|---|
| *Input Variables* | | | | | | | |
| CPK | 7.00 | 66.00 | 119.00 | 839.60 | 247.00 | 323,080 | 282,024 |
| CREA | 0.00 | 0.70 | 0.90 | 1.18 | 1.10 | 43.00 | 5,327 |
| CRP | 0.10 | 3.50 | 15.00 | 46.77 | 67.10 | 586.20 | 293,392 |
| LDH | 49.00 | 195.00 | 254.50 | 374.20 | 375.00 | 16,590 | 292,804 |
| UREA | 1.00 | 11.00 | 15.00 | 19.50 | 22.00 | 263.00 | 6,184 |
| aPTT | 2.70 | 27.70 | 30.30 | 33.08 | 34.20 | 150.00 | 169,294 |
| DDIMER | 50.50 | 216.00 | 372.00 | 844.20 | 768.00 | 25,135 | 294,077 |
| INR | 0.50 | 1.00 | 1.10 | 1.39 | 1.30 | 24.00 | 168,220 |
| HGB | 0.00 | 11.20 | 12.70 | 12.46 | 13.90 | 22.50 | 5,771 |
| LYM | 0.00 | 12.80 | 21.00 | 22.25 | 30.00 | 100.00 | 14,062 |
| NEUT | 0.00 | 58.70 | 68.60 | 67.80 | 78.20 | 100.00 | 14,062 |
| PLT | 5.00 | 185.00 | 233.00 | 243.50 | 287.00 | 2,947 | 6,462 |
| WBC | 0.00 | 6.30 | 8.10 | 9.13 | 10.70 | 632.10 | 6,037 |
| Age | 18.00 | 37.00 | 54.00 | 53.57 | 69.00 | 91.00 | 0.00 |
| *Output Variable - ED Admission* | | | | | | | |

An initial evaluation of the MIMIC dataset revealed missing values in nearly all the selected variables of interest. As a result, appropriate data preprocessing actions were undertaken to prepare the dataset for analysis and to enable comparison with the results of the study by Feretzakis et al. [49]. These actions followed a process consisting of four distinct steps.

Firstly, we thoroughly checked for extreme values using boxplots, as these can significantly affect model accuracy. An example of a boxplot created for this purpose is presented in Figure 1, focusing on the variable LHM (a complete set of figures for all examined variables is provided in the Appendix – Figure A1). Outliers were identified using Tukey's method [50], which calculates the Interquartile Range (IQR) as the distance between the 1st (Q1) and 3rd (Q3) quartiles, with boundaries at 1.5 times the IQR below Q1 and above Q3. Values outside these limits were considered extreme, ensuring model accuracy without compromising dataset integrity.

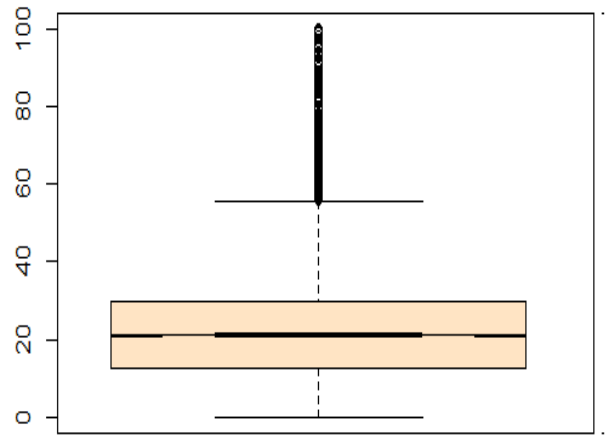

**Figure 1.** Boxplot of the LYM variable before data preparation in the MIMIC-IV dataset.

As a result of omitting extreme values, we have different descriptive Statistics of the MIMIC-IV data, highlighting in Table 3 with bold all the changes.

**Table 3.** Descriptive Statistics of Data Extracted from the MIMIC-IV Without Extreme Values

| Variable | Minimum Value | 1st Quartile | Median | Mean | 3rd Quartile | Maximum Value | Missing Values (NA's) |
|---|---|---|---|---|---|---|---|
| *Input Variables* | | | | | | | |
| CPK | 7.00 | **61.00** | 103.00 | 136.60 | 178.00 | 518.00 | 284,336 |
| CREA | **0.10** | 0.70 | 0.90 | **0.90** | 1.00 | 1.70 | 31,993 |
| CRP | 0.10 | **3.10** | 12.10 | 32.51 | 50.05 | 162.40 | 293,802 |
| LDH | 49.00 | **189.00** | 240.00 | 272.00 | 326.00 | 645.00 | 293,469 |
| UREA | 1.00 | 11.00 | 15.00 | **16.01** | 20.00 | 38.50 | 28,845 |
| aPTT | **18.00** | 27.40 | 30.00 | 30.54 | 33.00 | 43.94 | 178,670 |
| DDIMER | 50.50 | 204.00 | 325.50 | 437.60 | 564.00 | 1,596 | 294,810 |
| INR | **0.70** | 1.00 | 1.10 | **1.12** | 1.20 | 1.70 | 186,803 |
| HGB | **7.15** | 11.30 | 12.70 | 12.54 | 14.00 | 17.95 | 10,329 |
| LYM | 0.00 | **12.70** | 20.80 | 21.84 | 29.80 | 55.80 | 16,299 |
| NEUT | **29.50** | **59.00** | 68.70 | 68.22 | 78.30 | 100.00 | 16,126 |
| PLT | **32.00** | **184.00** | 230.00 | 233.80 | 281.00 | 440.00 | 17,286 |
| WBC | 0.00 | **6.20** | 8.00 | 8.39 | 10.20 | 17.27 | 18,579 |
| Age | 18.00 | 37.00 | 54.00 | **53.56** | 69.00 | 91.00 | 0.00 |
| *Output Variable - ED Admission* | | | | | | | |

Next, we examined the data for missing values, which, though minimal [51], were addressed to maintain dataset consistency. As a widely used and well-established approach in MIMIC literature for this purpose [52-53], we applied median substitution—a single imputation method—where missing values are replaced with the median of the feature, preserving the dataset's overall statistical properties. After this process, the mentioned descriptive statistics are reformed, according to Table 4.

**Table 4.** Descriptive Statistics of MIMIC-IV Data After Median Imputation

| Variable | Minimum Value | 1st Quartile | Median | Mean | 3rd Quartile | Maximum Value | Missing Values (NA's) |
|---|---|---|---|---|---|---|---|
| *Input Variables* | | | | | | | |
| CPK | 7.00 | **103.00** | 103.00 | **105.10** | 103.00 | 518.00 | - |
| CREA | 0.10 | 0.70 | 0.90 | **0.90** | 1.00 | 1.70 | - |
| CRP | 0.10 | **12.10** | 12.10 | **12.70** | 12.10 | 162.40 | - |
| LDH | 49.00 | **240.00** | 240.00 | **241.00** | 240.00 | 645.00 | - |
| UREA | 1.00 | 11.00 | 15.00 | **15.91** | 19.00 | 38.50 | - |
| aPTT | 18.00 | **30.00** | 30.00 | **30.22** | 30.00 | 43.95 | - |
| DDIMER | 50.50 | **325.50** | 325.50 | **328.6** | 325.50 | 1,596 | - |
| INR | 0.70 | **1.10** | 1.10 | **1.11** | 1.10 | 1.75 | - |
| HGB | 7.15 | 11.30 | 12.70 | **12.55** | 13.90 | 17.95 | - |
| LYM | 0.00 | **13.10** | 20.80 | **21.78** | 29.20 | 55.80 | - |
| NEUT | 29.50 | **59.60** | 68.70 | **68.25** | 77.70 | 100.00 | - |
| PLT | 32.00 | **187.00** | 230.00 | **233.60** | 277.00 | 440.00 | - |
| WBC | 0.00 | **6.30** | 8.00 | **8.36** | 10.00 | 17.27 | - |
| Age | 18.00 | 37.00 | 54.00 | 53.56 | 69.00 | 91.00 | - |
| *Output Variable - ED Admission* | | | | | | | |

Finally, we encoded the categorical data into a format suitable for ML analysis. For this, we used the widely adopted One-Hot Encoding method [46, 54], where each category of the original variable is transformed into a binary variable, assigned a value of 1 if it belongs to that category and 0 if it does not. This procedure specifically applied to the columns for gender, hospital attendance, triage disposition, and admission outcome.

After completing these steps, we obtained an initial overview of the MIMIC data, which can be compared side by side with the data from the study by Feretzakis et al. [49], as shown in Table 5.

**Table 5.** Statistics for Age, Gender and Admission.

|  | *Feretzakis et al. [49]* | *MIMIC-IV* |
|---|---|---|
|  | **Age** | |
| Mean | 61.85 | 53.56 |
| St. deviation | 20.82 | 19.88 |
| Range/IQR | 84/33 | 73/32 |
|  | ***Gender*** | |
| Male | 7,586 (54.22%) | 165,630 (54.63%) |
| Female | 6,405 (45.78%) | 137,503 (45.36%) |
|  | ***Admission*** | |
| Yes | 6,303 (45.05%) | 115,622 (38.14%) |
| No | 7,688 (54.95%) | 187,511 (61.86%) |
| **Total** | **13,991** | **303,133** |

As shown, the gender distribution is nearly identical across the two datasets: males constitute 54.22% (7,586 patients) in the Feretzakis dataset and 54.63% (165,630 patients) in MIMIC-IV, while females account for 45.78% (6,405 patients) and 45.36% (137,503 patients), respectively. Similarly, the age distribution aligns closely, with the mean age in the Feretzakis dataset slightly higher (61.85 vs. 53.56 years), but the interquartile ranges (33 and 32) are almost identical, reflecting comparable central age distributions. Lastly, while admission rates show some variation, they reflect broadly comparable structures. In the Feretzakis dataset, 45.05% (6,303 patients) were admitted, compared to 38.14% (115,622 patients) in MIMIC-IV. These differences likely reflect contextual variations between the two healthcare systems, but overall, the datasets remain structurally similar.

## 3.2. Algorithms Employed for Predictive Analysis

The algorithms applied to the MIMIC-IV data were like those used by Feretzakis et al. [49], including Random Forests (RF), K-Nearest Neighbor (KNN), Linear Discriminant Analysis (LDA) Recursive Partitioning and Regression Trees (RPART) and Support Vector Machines (svmRadial). These five algorithms are widely known in relevant literature.

Firstly, RF is one of the most usable algorithms for its ability to predict medical domain [55]. Despite its simplicity, it effectively avoids overfitting by leveraging averaging or voting mechanisms and applying the law of large numbers [56]. RF has been used in previous studies related to the prediction of ED admissions, including those by Araz et al. [57], Mowbray et al. [58], and Lucini et al. [59]. As a method, this algorithm creates a set of multiple decision trees to generate a prediction or outcome, with each tree constructed using a random subset of the dataset [60].

KNN is an algorithm notable for its simplicity and widespread popularity [61], often appearing in studies such as those by Abd-Elrazek et al. [62] and Ahmed et al. [63]. It is essentially a supervised learning algorithm that, given a specified k, calculates the Euclidean distances between the sample to be predicted and all training samples, sorts them, and selects the k-nearest neighbors [64].

Furthermore, LDA is a supervised learning method commonly used in the literature [65]. From a methodological perspective, the data is mapped to a lower-dimensional space in a way that maximizes the distance between class means while minimizing the variance within each class. This is accomplished by calculating the eigenvectors of the scatter matrix, which captures the interactions between the features and the classes [66].

Additionally, RPART is a decision tree algorithm that recursively splits the dataset into two subsets, using the features that most effectively reduce the heterogeneity of the outcome variables to determine each partition [67]. This algorithm is widely used in literature, as seen in studies such as Faisal et al. [68], Horwitz et al. [69], and Zelkowitz et al. [70].

Lastly, the fifth and final algorithm used in both studies is svmRadial. This algorithm identifies the hyperplane that best separates the data categories with the maximum margin, enabling improved classification of new data [71]. It is considered significant for its ability to enhance a model's predictive accuracy without overfitting the training set [72]. For this reason, it has been applied in predicting outcomes in EDs by researchers such as Tsoni et al. [20], Paliwal et al. [72], and Benevento et al. [73].

The algorithms were executed using RStudio [74], with the sole exception of the svmRadial algorithm, which was implemented in Python. To train the models, we utilized the "caret" package in R [75], specifically leveraging the train() function for streamlined model tuning and parameter selection. Additionally, for the RPART algorithm, we employed the "rpart" package [76].

To enhance model reliability and prevent overfitting [77], we applied 10-fold cross-validation. This method divided randomly the data into ten subsets, using nine for training and one for testing in each iteration, ensuring all subsets contributed to validation and resulting in a more generalized model with improved predictive accuracy [78].

*3.3. Evaluation Metrics Used*

In accordance with the study by Feretzakis et al. [49], three evaluation metrics were used to present the results of predicting ED admissions based on MIMIC-IV data: Sensitivity (or Recall), Specificity, and AUC-ROC.

The AUC-ROC is a very popular and quantitative measure of the model's performance [79,80]. It provides the probability of correctly classifying observations, where a value of 0 indicates misclassification into opposite classes, 1 indicates perfect classification, and 0.5 reflects random predictions [81].

The Sensitivity index is also a commonly used performance metric [36, 82-84]. Especially, it measures the model's ability to accurately identify positive classes and is defined by the following formula 1. For this metric, a value closer to 1 indicates that the algorithm has more accurately estimated the positive classes [82].

$$\text{Sensitivity} = \frac{TP}{TP+FN} \quad (1)$$

*where*

- TP (True Positive) represents the number of positive class observations that were correctly classified
- FN (False Negative) represents the number of positive class observations that were incorrectly classified as belonging to the negative class

Specificity index is also a measure that is used in the relative literature, commonly with sensitivity index [36, 84]. It measures the model's ability to estimate correctly the negative classes, with as the value is near to 1 the more the algorithm estimated the more correctly the negative glasses. The formula for calculating this index is given by equation 2 [85].

$$\text{Specificity} = \frac{TN}{TN+FP} \quad (2)$$

*where*

- TN (True Negative) represents the number of negative class observations that were correctly classified

- FP (False Positive) represents the number of negative class observations that were incorrectly classified as belonging to the positive class

## 4. Results

Table 6 presents the comparative results of the five algorithms examined across the two different analysis cases, noting that the MIMIC-IV database includes a significantly larger number of ED visits than the study by Feretzakis et al. [49].

**Table 6.** Comparative results of four algorithms across the two study analyses.

|  | *Feretzakis et al. [49]* | *MIMIC-IV* |
|---|---|---|
| | *AUC-ROC* | |
| LDA | 0.7834 | 0.9387 |
| KNN | 0.7307 | 0.7112 |
| **RF** | **0.8054** | **0.9999** |
| RPART | 0.6989 | 0.9092 |
| svmRadial | 0.7961 | 0.7640 |
| | *Sensitivity Index* | |
| LDA | **0.7168** | 0.9809 |
| KNN | 0.6778 | 0.6389 |
| RF | 0.6969 | **0.9997** |
| RPART | 0.5834 | 0.9849 |
| svmRadial | 0.7013 | 0.6980 |
| | *Specificity Index* | |
| LDA | 0.7184 | 0.8356 |
| KNN | 0.6800 | 0.7835 |
| **RF** | **0.7757** | **0.9999** |
| RPART | 0.7617 | 0.8236 |
| svmRadial | 0.7687 | 0.4680 |

According to the results, there are differences between the two analyses, but also significant similarities across the various methods used to evaluate the predictive performance of the algorithms under consideration.

Regarding the AUC-ROC, all algorithms demonstrated improved performance when applied to the larger and more diverse MIMIC-IV dataset, with consistent trends observed across analyses. RF achieved the highest AUC-ROC values in both studies, starting with good predictive accuracy in the Feretzakis et al. [49] study (0.7813) and reaching near-perfect classification in MIMIC-IV (0.9999). LDA followed as the next strongest performer, improving from a moderate AUC-ROC of 0.7569 in Feretzakis et al. [49] to an impressive 0.9386 in MIMIC-IV. Similarly, svmRadial showed strong and consistent performance, with an AUC-ROC of 0.7961 in the Feretzakis et al. [49] study, slightly decreasing to 0.764 in MIMIC-IV, indicating it maintained relatively high predictive accuracy. RPART displayed moderate performance in Feretzakis et al. [49] (AUC-ROC: 0.6995) but improved substantially in MIMIC-IV, achieving 0.9093. Finally, KNN had the lowest AUC-ROC values in both datasets (0.6832 in Feretzakis et al. [49] and 0.7106 in MIMIC-IV), showing only modest benefit from the larger dataset.

Following the trends observed with the AUC-ROC, the Sensitivity Index further highlights the impact of the MIMIC-IV dataset on algorithm performance. RF once again led the results, with sensitivity increasing from 0.6969 in the Feretzakis et al. [49] study to an almost perfect 0.9997 in MIMIC-IV, showcasing its exceptional ability to detect positive cases. Similarly, LDA demonstrated a substantial improvement, rising from 0.7168 in Feretzakis et al. [49] to 0.9809 in MIMIC-IV, mirroring its strong gains in AUC-ROC. RPART followed a comparable trend, with sensitivity improving significantly from 0.5834 in Feretzakis et al. [49] to 0.9849 in MIMIC-IV, reinforcing its capacity to adapt to more extensive and diverse datasets. On the other hand, svmRadial showed minimal variation, maintaining sensitivity values of 0.7013 in Feretzakis et al. [49] and 0.698 in MIMIC-IV, indicating its stable but limited adaptability. In contrast, KNN's sensitivity declined from 0.6778 in Feretzakis et al. [49] to 0.6389 in MIMIC-IV, further reflecting its weaker performance compared to other models.

Building on the improvements seen in AUC-ROC and Sensitivity Index, the Specificity Index also reflected notable changes when the algorithms were applied to the MIMIC-IV dataset. RF again led the results, with specificity rising from 0.7757 in the Feretzakis et al. [49] study to a near-perfect 0.9999 in MIMIC-IV, indicating its strong ability to correctly identify negative cases. LDA showed a moderate improvement, increasing from 0.7184 in Feretzakis et al. [49] to 0.8356 in MIMIC-IV, demonstrating its enhanced performance with the larger dataset. RPART showed a slight improvement, specifically rising from 0.7617 in Feretzakis et al. [49] to 0.8236 in MIMIC-IV, maintaining a solid performance. Similarly, KNN improved from 0.6800 in Feretzakis et al. [49] to 0.7835 in MIMIC-IV, reflecting a modest increase in its ability to identify negative cases. In contrast, svmRadial exhibited a significant drop in specificity, decreasing from 0.7687 in Feretzakis et al. [49] to 0.468 in MIMIC-IV, highlighting a notable decline in its performance with the larger dataset.

Regarding the best-performing model, we additionally present two figures: the ROC curve and the confusion matrix. As shown, the RF model demonstrates exceptional classification performance, achieving near-perfect discrimination between positive and negative cases. The ROC curve highlights its strong predictive ability, while the confusion matrix confirms its high accuracy in both sensitivity and specificity. These results further validate the model's robustness and its improved performance when applied to a larger and more diverse dataset.

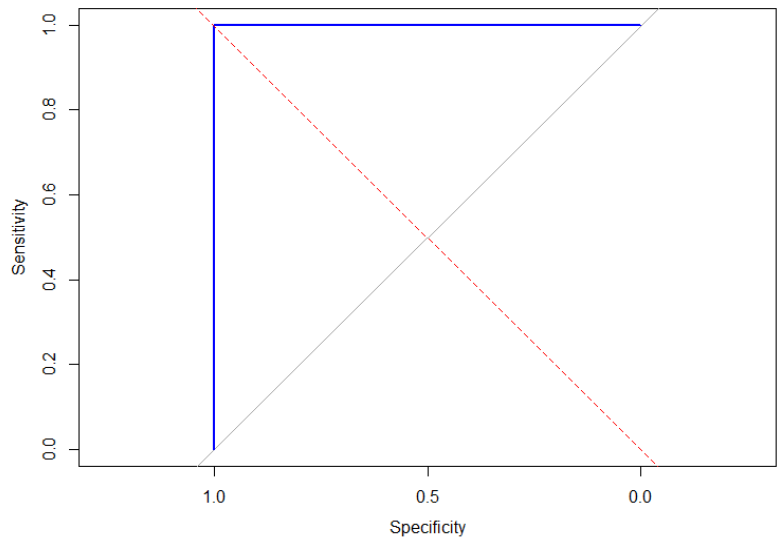

**Figure 2.** ROC Curve for the RF Model.

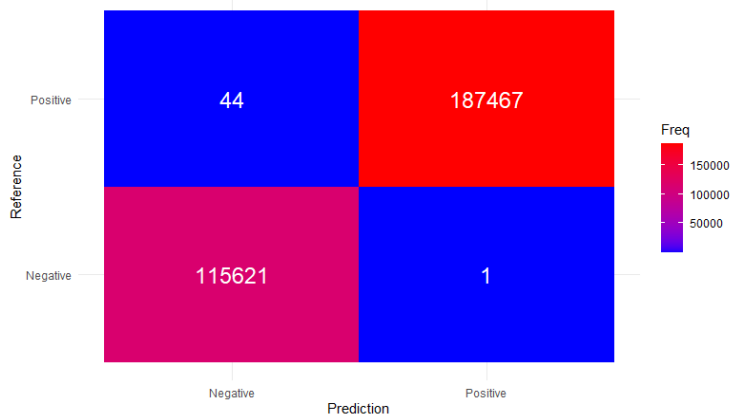

**Figure 3.** Confusion Matrix Heatmap of the RF model.

## 5. Discussion

The aim of this study was to assess the extent to which MIMIC-IV, as a large-scale dataset, can validate ED admission prediction models previously evaluated on a smaller local dataset. With numerous ongoing studies worldwide relying on small local datasets, identifying a robust dataset capable of validating these findings after appropriate preprocessing has become a critical challenge. Such validation would enable intensive care units to more effectively manage the growing influx of high-intensity patients, regardless of the healthcare settings in which they operate.

In this context, we applied MIMIC-IV in the same analytical framework used by Feretzakis et al. [49] to study the prediction of ED admissions based on local data from a Greek tertiary public hospital. A critical step in this process was the meticulous preprocessing of the MIMIC-IV data to ensure that the demographic characteristics of the two datasets were closely aligned, enabling a meaningful and accurate comparison.

After completing the necessary preprocessing steps and executing the algorithms, we reached valuable conclusions regarding the effectiveness of the MIMIC-IV database as a tool for validating the results of smaller studies. By applying five commonly used algorithms—LDA, KNN, RF, RPART, and svmRadial—we obtained critical insights that align with the findings of the reference study. Notably, these models are designed to assist clinicians in making informed decisions rather than replace their expertise, enhancing efficiency and consistency in ED admissions. In our analysis, we utilized 10-fold cross-validation, which divides the data into ten subsets, using nine for training and one for testing in each iteration. This method ensures that all data points are used for both training and validation, contributing to a more generalized model and improving predictive accuracy. Therefore, we did not perform separate training and testing phases but instead relied on this cross-validation process to assess the model's performance.

The first critical finding of this study was that, consistent with the research by Feretzakis et al. [49], the RF model outperformed other algorithms in terms of predictive ability, underscoring its robustness and flexibility in handling complex, heterogeneous healthcare data. Specifically, this model achieved near-perfect classification with MIMIC-IV (AUC-ROC 0.9999), demonstrating a significant performance difference when compared to the results reported in the Greek local dataset study (AUC-ROC 0.7813). This result suggests that the comprehensive and detailed data available in MIMIC-IV enables RF to more effectively capture patterns and interactions related to ED admissions, making it an ideal model for predictive tasks involving large, complex datasets.

Furthermore, we reached the same conclusion with two other evaluation metrics used in our study, the Sensitivity and Specificity indices. In both cases, the RF algorithm stood out compared to the other models, while the values of the other algorithms were generally improved by utilizing the larger MIMIC-IV database.

However, despite our promising findings, it is essential to acknowledge some possible limitations of our study. The conclusions regarding MIMIC-IV are based on a local dataset with almost identical descriptive statistics, achieved through data preprocessing. So, in another local dataset, it may be more difficult to obtain comparable results, potentially leading to varying outcomes or no improvement in predictive accuracy for ED admissions. Effective validation using MIMIC-IV requires careful alignment of variables for accurate comparison and proper preprocessing of missing values and outliers, as data quality issues may introduce biases that compromise model accuracy. Additionally, the extremely high performance of the RF model (AUC-ROC = 0.9999, sensitivity = 0.9997, specificity = 0.9999) may suggest a potential risk of overfitting. While these results indicate excellent classification accuracy, such near-perfect scores may imply that the model has learned dataset-specific patterns that do not generalize well to different ED environments. Overfitting is a critical concern, as it can lead to misleadingly optimistic performance metrics that fail when applied to new, unseen data. Another relevant limitation of our study is the lack of an analysis of model explainability, such as using SHAP values, which could provide more transparency into the decision-making process of the models. While the focus of this study was primarily on model performance, we acknowledge that explainability is crucial—especially in high-stakes environments like EDs. Understanding which features influence predictions could increase trust in the model and facilitate its adoption by healthcare professionals.

Moreover, a single dataset may not be sufficient for generalizing the results. Finding additional local datasets for analysis is a time-consuming and demanding process, as it requires obtaining the appropriate permissions for data access and use in research. Furthermore, although we used five of the most widely applied prediction algorithms in ED literature, as suggested by the comparative study by Feretzakis et al.

[49], there are still several notable algorithms in the literature, such as decision trees [20, 57, 86] and gradient boosted machines [86], which could provide further insights into the utility of MIMIC-IV for this specific purpose. This reflects the importance of continuously exploring new techniques and refining existing ones, especially as ED environments evolve.

One notable challenge in the practical implementation of predictive models for EDs is ensuring healthcare professionals are adequately trained to interpret predictive analytics, particularly in high-pressure environments [87-88]. The ability to trust and act on predictions may be hindered by the model's complexity or lack of transparency, an aspect that needs careful consideration when deploying models in real-world settings. Additionally, the model's performance might vary depending on the clinical context, making it critical to assess its usability in various ED environments.

These considerations provide strong motivation for future research. Further validation studies using the MIMIC-IV database could help confirm or challenge the conclusions of the present study. Validation tests across studies that utilize local databases from various countries, with differing sample compositions and algorithms, would be particularly valuable. While this study followed the algorithms used in the local data study by Feretzakis et al. for consistency in comparison, future research could explore additional machine learning models such as XGBoost and LightGBM to assess their potential benefits. Similarly, expanding the evaluation metrics to include classification accuracy and F1-score could provide a more comprehensive assessment of model performance. This process could yield essential insights into the development of a generalized prediction model capable of addressing global health challenges. Such a model could support the efficient functioning of EDs worldwide, contributing to the overall safeguarding of public health.

## 6. Conclusion

In conclusion, this study evaluates the efficacy of the MIMIC-IV dataset as a robust resource for refining and improving ED admission predictive models previously tested on smaller local datasets. The superior performance of the RF algorithm, particularly with MIMIC-IV data, highlights the potential of such models to assist doctors by providing data-driven insights that complement their clinical expertise, rather than substituting their judgment. Integrating MIMIC-IV or similar datasets with local healthcare data could significantly enhance predictive accuracy, supporting healthcare providers and policymakers in making data-driven decisions to alleviate ED overcrowding and improve patient outcomes. Future research should further explore cross-dataset validations and model adaptations using the MIMIC-IV database to confirm improvements in predictive accuracy, ensuring that these models function effectively as decision-support tools that enhance, rather than replace, physician judgment in diverse healthcare settings. Additionally, comparative validation studies can serve as a valuable sanity check by identifying high-level similarities and differences between communities, thereby enhancing the agility of interdisciplinary teams of medical professionals and data scientists working collaboratively on local data. Finally, a promising direction for future research is the development of targeted educational programs that equip healthcare professionals with the skills needed to understand and effectively use predictive models, such as those derived from the MIMIC-IV database. Pedagogical frameworks should be implemented to integrate ML tools into everyday clinical practice. This can be achieved through the creation of interactive, multidisciplinary training modules that offer hands-on experience in interpreting model predictions, understanding their limitations, and applying them in decision-making processes.

**Author Contributions**: Conceptualization, D.K. and F.M.; methodology, D.K. and F.M.; software, F.M. and G.F.; validation, F.M.; formal analysis, F.M.; investigation, F.M.; resources, A.S., V.K., L.Tz., N.T., G.F. and F.M.; data curation, A.S., V.K., L.Tz., N.T., G.F. and F.M.; writing—original draft preparation, L.Tr. and F.M.; writing—review and editing, A.S., V.K., L.Tz., N.T., E.P., G.F., D.K., V.V.; supervision, D.K. and V.V.; project administration, D.K., L.Tr. and V.V. All authors have read and agreed to the published version of the manuscript.

**Funding:** This research received no external funding.

**Data Availability Statement:** The data presented in this study is only contained in the article itself.

**Conflicts of Interest:** The authors declare no conflicts of interest.

# Appendix A

Table A1. SQL Queries Used to Obtain ED Patient and Lab Test Data.

```sql
-- Creates 'table_subjectinEDandlab' with ED patients and their lab test data.
-- It combines data from the edstays, patients, and labevents tables, linking them via the subject_id.
-- The lab orders are filtered to ensure they occurred between the patient's ED entry and exit times.

CREATE TABLE table_subjectinEDandlab AS
SELECT
edstays.subject_id,patients.gender, patients.anchor_age as age,edstays.hadm_id,
edstays.arrival_transport, edstays.disposition,edstays.intime,
edstays.outtime,labevents.charttime,labevents.specimen_id,itemid,value,valuenum,valueuom,ref_range_lower,ref_range_upper,flag
FROM labevents inner join edstays on
labevents.subject_id=edstays.subject_id and labevents.charttime between edstays.intime and edstays.outtime
inner join patients on patients.subject_id=edstays.subject_id;

-- Mapped biochemical markers to test codes in the MIMIC d_labitems table, with input from pathologists.
-- D-Dimer values had different units: ng/mL FEU and ng/mL DDU.
-- Updated the database to standardize the values using DDU = FEU / 2.
-- Create a new column 'valuenum_new'

ALTER TABLE table_subjectinEDandlab
ADD valuenum_new REAL;

-- Copy D-Dimer values to the new column

UPDATE table_subjectinEDandlab
SET valuenum_new = valuenum
WHERE itemid = "50915" OR itemid = "51196";

-- Convert FEU values to DDU for consistency

UPDATE table_subjectinEDandlab
SET valuenum_new = (valuenum / 2)
WHERE itemid = "50915" AND valueuom = "ng/mL FEU";

-- Querying table_subjectinEDandlab to retain specified biochemical indicators and average values for repeated tests.
-- Created a new table 'table_group_a'.

CREATE TABLE table_group_a AS
SELECT subject_id,gender,anchor_age,hadm_id,arrival_transport, disposition,intime,
outtime,charttime,specimen_id,
FORMAT(avg(case when itemid =="50910" then valuenum end),2) as [CPK],
FORMAT(avg(case when itemid =="52024" or itemid=="50912" or itemid=="52546" then valuenum end),2) as [CREA],
FORMAT(avg(case when itemid =="50889" or itemid=="51652" then valuenum end),2) as [CRP],
FORMAT(avg(case when itemid =="50954" then valuenum end),2) as [LDH],
FORMAT(avg(case when itemid =="51006" or itemid=="52647" then valuenum end),2) as [UREA],
FORMAT(avg(case when itemid =="51275" then valuenum end),2) as [aPTT],
FORMAT(avg(case when itemid =="50915" or itemid =="51196" then valuenum_new end),2) as [DDIMER],
FORMAT(avg(case when itemid =="51237" then valuenum end),2) as [INR],
FORMAT(avg(case when itemid =="50811" or itemid=="51222" then valuenum end),2) as [HGB],
FORMAT(avg(case when itemid =="51245" or itemid=="51244" then valuenum end),2) as [LYM],
FORMAT(avg(case when itemid =="51256" then valuenum end),2) as [NEUT],
FORMAT(avg(case when itemid =="51265" then valuenum end),2) as [PLT],
FORMAT(avg(case when itemid =="51300" or itemid=="51301" then valuenum end),2) as [WBC]
FROM table_subjectinEDandlab
group by subject_id, intime
order by subject_id, intime;
```

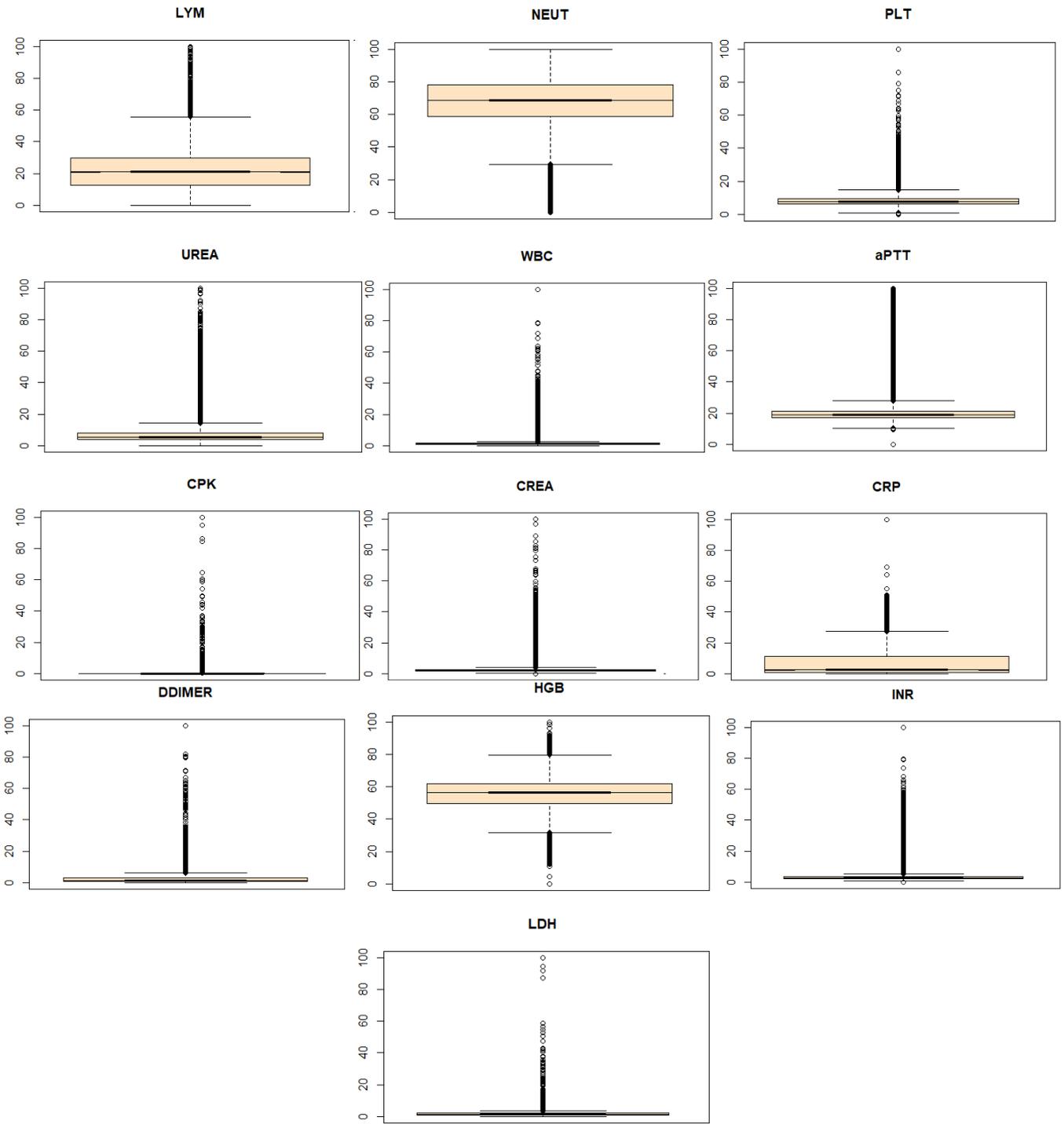

**Figure A1.** Boxplot of the examined variables before data preparation in the MIMIC-IV dataset.